\documentclass[10pt,twocolumn,letterpaper]{article}

\usepackage{subfig}

\usepackage{iccv}
\usepackage{times}
\usepackage{epsfig}
\usepackage{graphicx}
\usepackage{amsmath}
\usepackage{amssymb}

\usepackage{multirow}
\usepackage{booktabs}
\usepackage{xcolor}

\newcommand\scalemath[2]{\scalebox{#1}{\mbox{\ensuremath{\displaystyle #2}}}}

\usepackage[pagebackref=true,breaklinks=true,letterpaper=true,colorlinks,bookmarks=false]{hyperref}

\iccvfinalcopy %

\def\iccvPaperID{****} %

\begin{document}

\title{Graph-to-3D: End-to-End Generation and Manipulation of 3D Scenes Using Scene Graphs}

\author{Helisa Dhamo $^{1,}$\thanks{The first two authors contributed equally to this work} \hspace{0.87cm}
Fabian Manhardt $^{2,}$\footnotemark[1] \hspace{0.87cm}
Nassir Navab $^{1}$ \hspace{0.87cm}
Federico Tombari $^{1,2}$
\and $^{1}$ Technische Universit\"at M\"unchen
\hspace{0.25cm}
$^{2}$ Google
}

\maketitle

\begin{abstract}
   Controllable scene synthesis consists of generating 3D information that satisfy underlying specifications. Thereby, these specifications should be abstract, i.e. allowing easy user interaction, whilst providing enough interface for detailed control. Scene graphs are representations of a scene, composed of objects (nodes) and inter-object relationships (edges), proven to be particularly suited for this task, as they allow for semantic control on the generated content. Previous works tackling this task often rely on synthetic data, and retrieve object meshes, which naturally limits the generation capabilities. To circumvent this issue, we instead propose the first work that directly generates shapes from a scene graph in an end-to-end manner. In addition, we show that the same model supports scene modification, using the respective scene graph as interface. Leveraging Graph Convolutional Networks (GCN) we train a variational Auto-Encoder on top of the object and edge categories, as well as 3D shapes and scene layouts, allowing latter sampling of new scenes and shapes.
\end{abstract}

\section{Introduction}

Scene content generation, including 3D object shapes, images and 3D scenes is of high interest in computer vision. Applications involve helping the work of designers through automatically generated intermediate results, as well as understanding and modeling scenes, in terms of, \eg, object constellations an co-occurrences. Furthermore, conditional synthesis allows for a more controllable content generation, since users can specify which image or 3D model they want to let appear in the generated scene. Common conditions involve text descriptions~\cite{Zhang_2017_ICCV}, %
semantic maps~\cite{wang2018pix2pixHD} and scene graphs. Thereby, scene graphs have recently shown to offer a suitable interface for controllable synthesis and manipulation \cite{johnson2018image,Dhamo2020cvpr,Luo_2020_CVPR}, enabling semantic control on the generated scene, even for complex scenes. Compared to dense semantic maps, scene graph structures are more high-level and explicit, simplifying the interaction with the user. Moreover, they enable controlling the semantic relation between entities, which is often not captured in a semantic map. %

\begin{figure}[t!]
\begin{center}
   \includegraphics[width=0.95\linewidth]{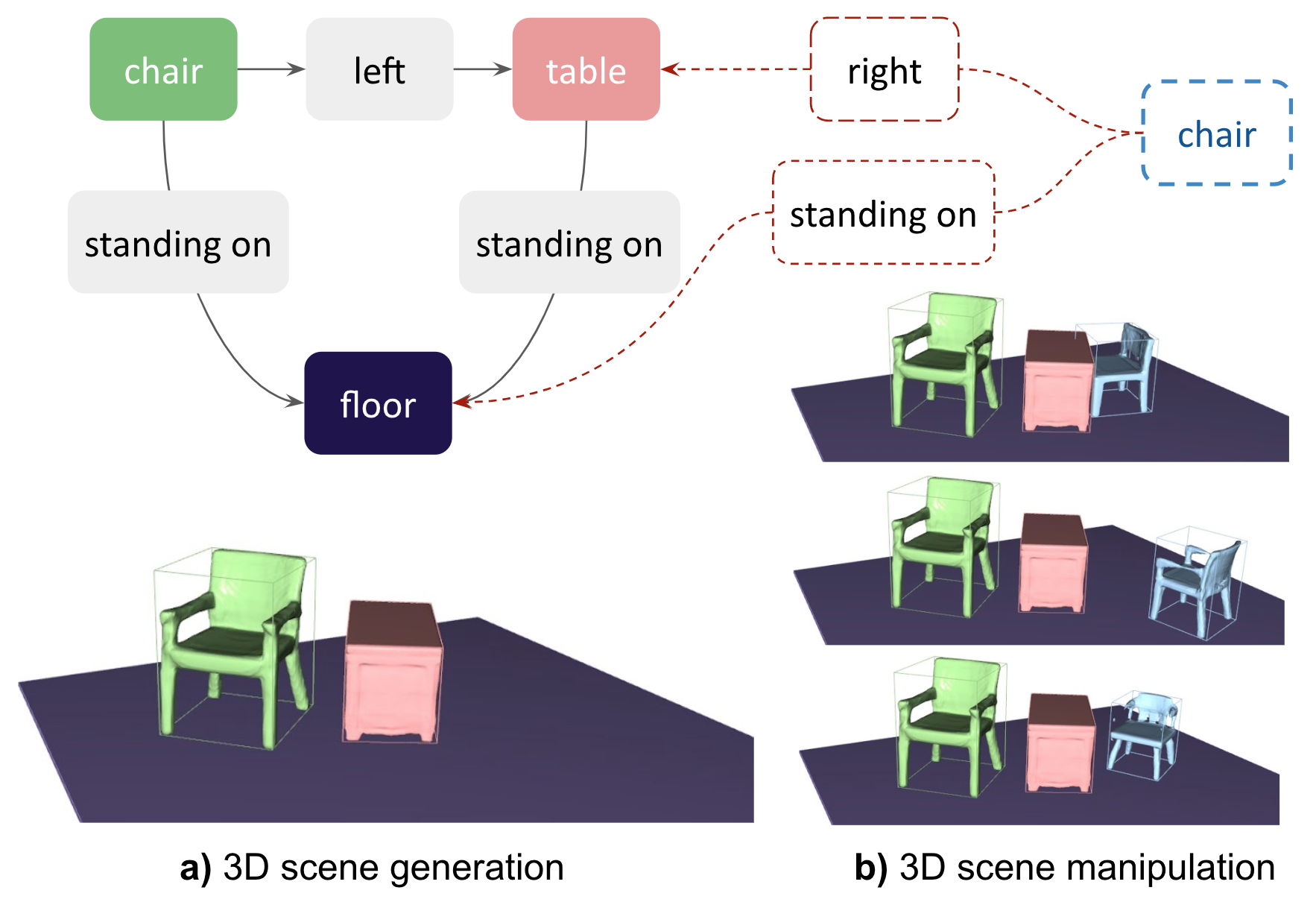}
\end{center}
   \caption{%
   a) \textit{Scene generation}: given a scene graph (top, solid lines), Graph-to-3D generates a 3D scene consistent with it. b) \textit{Scene manipulation}: given a 3D scene and an edited graph (top, solid+dotted lines), Graph-to-3D is able to generate a varied set of 3D scenes adjusted according to the graph manipulation.}
\label{fig:teaser}
\end{figure}

While there are a lot of methods for scene graph inference from images~\cite{xu2017scenegraph,newell2017pixels} as well as the reverse problem~\cite{johnson2018image,ashual2019specifying}, in the 3D domain, only a few works on scene graph prediction from 3D data have been very recently presented~\cite{3DSSG2020,Wu2021}. With this work, we thus attempt to fill this gap by proposing a method for end-to-end generation of 3D scenes from scene graphs.
A few recent works investigate the problem of scene layout generation from scene graphs~\cite{Wang2019PlanITPA,Luo_2020_CVPR}, thereby predicting a set of top-view object occupancy regions or 3D bounding boxes. To construct a 3D scene from this layout, these methods typically rely on retrieval from a database. %
On the contrary, we employ a fully generative model that is able to synthesize novel context-aware 3D shapes for the scene. Though retrieval leads to good quality results, shape generation is an emerging alternative as it allows further costumizability via interpolation at the object level~\cite{groueix2018atlas} and part level~\cite{Mo2019StructureNet}. Further, retrieval works can achieve at best (sub-)~linear complexity for time and space \wrt database size. 
Our method essentially predicts object-level 3D bounding boxes together with appropriate 3D shapes, which are then combined to create a full 3D scene (Figure~\ref{fig:teaser}, left). Leveraging Graph Convolutional Networks (GCNs) we learn a variational Auto-Encoder on top of scene graphs, 3D shapes and scene layouts, enabling latter sampling of novel scenes. Additionally, we employ a graph manipulation network to enable changes, such as adding new objects as well as changing object relationships, while maintaining the rest of the scene (Figure~\ref{fig:teaser}, right). To model the one-to-many problem of label to object, we introduce a novel relationship discriminator on 3D bounding boxes that does not limit the space of valid outputs to the annotated box. %

To avoid inducing any human bias, we want to learn 3D scene prediction from real data. However, these real datasets, such as 3RScan typically present additional limitations, such as information holes and, oftentimes, lack of annotations for the canonical object pose. We overcome the former limitation by refining the ground truth 3D boxes based on the semantic relationships from 3DSSG~\cite{3DSSG2020}. For the latter, we extract oriented 3D bounding boxes and annotate the front side of each object, using a combination of class-level rules and manual annotations. We release these annotations as well as the source code on our project page\footnote{Project page: \url{https://he-dhamo.github.io/Graphto3D/}}.

Our contributions can be summarized as: i) We propose the first fully learned method for generating a 3D scene from a scene graph. Therefore, we use a novel model for shared layout and shape generation. ii) We also adopt this generative model to simultaneously allow for scene manipulation. iii) We introduce a relationship discriminator loss which is better suited than reconstruction losses due to the one-to-many problem of box inference from class labels. iv) We label 3RScan with canonical object poses.

We evaluate our proposed method on 3DSSG~\cite{3DSSG2020}, a large-scale real 3D dataset based on 3RScan~\cite{Wald2019RIO} that contains semantic scene graphs. Thereby, we evaluate on common aspects of scene generation and manipulation, such as quality, diversity and fulfillment of relational constrains, showing compelling results, as well as an advantage of sharing layout and shape features for both tasks. 

\section{Related work}

\paragraph{Scene graphs and images}

Scene graphs~\cite{johnson15,krishna2017visual} refer to a representation that provides a semantic description for a given image. Whereas nodes depict scene entities (objects), edges represent the relationships between them. 
A line of works focuses on scene graph prediction from images ~\cite{xu2017scenegraph,herzig2018mapping,qi2018attentive,zellers2018neural,li2017scene,yang2018graph,li2018factorizable,newell2017pixels}. Other work explore scene graphs for tasks such as image retrieval~\cite{johnson15}, image generation~\cite{johnson2018image,ashual2019specifying} and manipulation~\cite{Dhamo2020cvpr}. 

\paragraph{Scene graphs in 3D} %
The 3D computer vision and graphics communities have proposed a diverse set of scene graph representations and related structures. Scenes are often represented through a hierarchical tree, where the leaves are typically objects and the intermediate nodes form (functional) scene entities~\cite{Li2018grains,Liu:2014:CCS,NIPS2011_4236}. Armeni \etal~\cite{armeni_iccv19} propose a hierarchical mapping of 3D models of large spaces in four layers: camera, object, room and building. %
Wald \etal~\cite{3DSSG2020} introduce 3DSSG, a large scale dataset with dense semantic graph annotations. %
These graph representations are utilized to explore tasks related to scene comparison~\cite{Fisher11characterizingstructural}, scene graph prediction \cite{3DSSG2020}, 2D-3D scene retrieval \cite{3DSSG2020}, layout generation~\cite{Luo_2020_CVPR}, object type predictions in query locations~\cite{zhou2019scenegraphnet}, as well as to improve 3D object detection \cite{shi2019hierarchy}.

\vspace{5pt}
\noindent \textbf{3D scene and layout generation} A line of works generates 3D scenes conditioned on images~\cite{factored3dTulsiani17,Nie_2020_CVPR}. %
Jiang \etal~\cite{Jiangijcv18} use probabilistic grammar to control scene synthesis. Other works, more related to ours, incorporate graph structures. StructureNet~\cite{Mo2019StructureNet} explores an object-level hierarchical graph, to generate shapes in a part-aware model. Ma \etal~\cite{Ma2018language} convert text to a scene graph with pairwise and group relationships, to progressively retrieve sub-scenes for 3D synthesis. While generative methods were recently explored for layouts of different types~\cite{Jyothi_2019_ICCV}, some methods focus on generating scene layouts. GRAINS~\cite{Li2018grains} explore hierarchical graphs to generate 3D scenes, using a recursive VAE that generates a layout, followed by object retrieval. 
Luo \etal~\cite{Luo_2020_CVPR} generate a 3D scene layout conditioned on a scene graph, combined with a rendering approach to improve image generation. Other works use deep priors~\cite{wang2018deep} or relational graphs~\cite{Wang2019PlanITPA} to learn object occupancy in the top-view of indoor scenes.

Different from our work, these works either explore images as final output, use 3D models based on retrieval, or operate on synthetic scenes. Hence, these methods can either not fully explain the actual 3D scene or are not capable of generating context-aware real compositions.
\section{Data preparation}
\label{sec:preparation}
Our approach is built on top of 3DSSG~\cite{3DSSG2020}, a scene graph extension of 3RScan~\cite{Wald2019RIO}, which is a large-scale indoor dataset with $\sim$1.4k real 3D scans. 3RScan does not contain canonical poses for objects, which is essential to learning object pose and shape as well as many other tasks. 

Therefore, we implemented a fast semi-automatic annotation pipeline to obtain canonical tight bounding boxes per instance. As most objects are supported by a horizontal surface, we model the oriented boxes with 7 degrees-of-freedom (7DoF), \ie 3 for size, 3 for translation as well as 1 for the rotation around the z-axis. %
Since the oriented bounding box should fully enclose the object whilst possessing minimal volume, we use volume as criteria to optimize the rotational parameter. First, for each object we extract the point set $p$. Then, we gradually rotate the points along the z-axis using angles $\alpha$ in the range $[0, 90[$ degrees, with a step of 1 degree, $p_t = R(\alpha) p$. At each step, we extract the axis-aligned bounding box from the transformed point set $p_t$, by simply computing the extrema along each axis. We estimate the area of the 2D bounding box in bird's eye view, after applying an orthogonal projection onto the ground plane. %
We then label the rotation $\hat{\alpha}$ having the smallest box top-down view area (\textit{c.f.} supplementary material). From this box we extract the final box parameters: width $w$, length $l$ and height $h$, rotation $\hat{\alpha}$ as well as centroid $(c_x, c_y, c_z)$.

The extracted bounding box remains still ambiguous, as there are always four possible solutions regarding the facing direction. Hence, for objects with two or more vertical axes of symmetry, such as tables, we automatically define as front the largest size component (in line with ShapeNet~\cite{chang2015shapenet}). For all other objects such as chair or sofa the facing direction is annotated manually (4.3k instances in total). %

As 3D boxes are obtained from the object point clouds, we observe misalignments due to impartial scans. Objects are oftentimes not touching their supporting structures, \eg chair with missing legs leads to a "flying" box detached from the floor. We thus detect inconsistencies using the support relationships from~\cite{3DSSG2020}. If an object has a distance of more that $10cm$ from its support, we fix the respective 3D box, such that it reaches the upper level of the parent object. For planar support such as floor, we employ RANSAC~\cite{fischler1981random} to fit a plane in a neighbourhood around the object and extend the object box so that it touches the fitted plane.

\section{Methodology}

In this work we propose a novel method for generating full 3D scenes from a given scene graph in a fully learned and end-to-end fashion. In particular, given a scene graph $G=(\mathcal{O},\mathcal{R})$, where nodes $ o_i \in \mathcal{O}$ are semantic object labels and edges $r_{ij} \in \mathcal{R}$ are semantic relationship labels with $i \in \{1,...,N\}$ and $j \in \{1,...,N\}$, we generate a corresponding 3D scene $S$. Throughout this paper we will utilize the notation $n_i \in \mathcal{N}$ to refer to nodes more generally. We represent the 3D scene $S=(\mathcal{B}, \mathcal{S})$ as a set of per-object bounding boxes $\mathcal{B} = \{b_0,...,b_N\}$ and shapes $\mathcal{S} = \{s_0,...,s_N\}$. Inspired by~\cite{Luo_2020_CVPR} on layout generation for image synthesis, we base our model on a variational scene graph Auto-Encoder. However, whereas \cite{Luo_2020_CVPR} relies on shape retrieval, we jointly learn layouts and shapes via a shared latent embedding, as these are two inherently cohesive tasks strongly supporting each other. Moreover, we enable scene manipulation in the \emph{same} learned model, using the scene graph as interface. In particular, given a scene together with its scene graph, changes can be applied to the scene, by interacting with the graph, such as adding new nodes or changing relationships. We do not need to learn object removal as this can be easily achieved by dismissing the corresponding box and shape for the given node.

\begin{figure*}[t]
\begin{center}
   \includegraphics[width=0.95\linewidth]{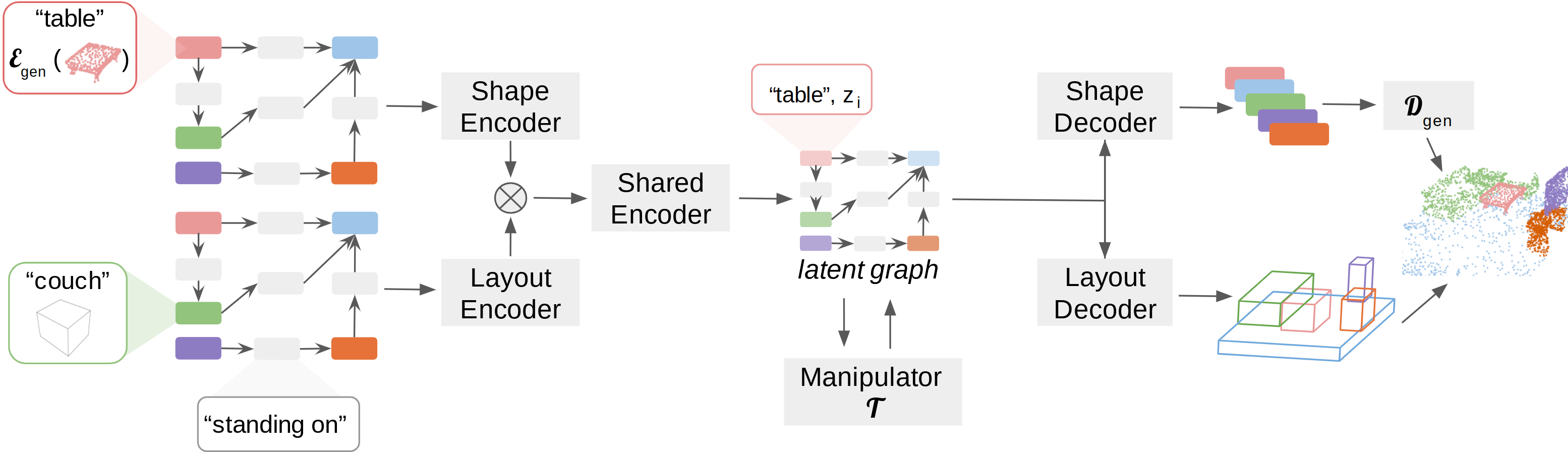}
\end{center}
   \caption{\textbf{Graph-to-3D pipeline}. Given a scene graph we generate a set of bounding boxes and object shapes. We employ a graph-based variational Auto-Encoder with two parallel GCN encoders sharing latent box and shape information through a shared encoder module. Given a sample from the learned underlying distribution the final 3D scene is obtained via combining the predictions from individual GCN decoders for 3D boxes and shapes. We further use a GCN manipulator for on the fly incorporation of user modifications to the scene graph.}
\label{fig:method}
\end{figure*}

The overall architecture is demonstrated in Figure~\ref{fig:method}. We first process scene graphs through a layout $\mathcal{E}_\mathrm{layout}$ and shape $\mathcal{E}_\mathrm{shape}$ encoder, section~\ref{sec:layout_en}. We then employ a shared encoder $\mathcal{E}_\mathrm{shared}$ which combines features from $\mathcal{E}_\mathrm{layout}$ and $\mathcal{E}_\mathrm{shape}$, section \ref{sec:shared}. This shared embedding is further fed to a shape $\mathcal{D}_\mathrm{shape}$ and layout $\mathcal{D}_\mathrm{layout}$ decoder to obtain the final scene. Finally, we use a modification network $\mathcal{T}$ (section~\ref{sec:interaction}) to enable the model the incorporation of changes in the scene while preserving the unchanged parts.

\subsection{Graph Convolutional Network}

At the heart of each building block in our model lies a Graph Convolutional Network (GCN) with residual layers~\cite{li2019deepgcns}, which enables information flow between the connected objects of the graph. Each layer $l_g$ of the GCN operates on directed relationships triplets (\emph{out -- p -- in}) and consists of three steps. First, each triplet $ij$ is fed in a Multi-Layer Perceptron (MLP) $g_1(\cdot)$ for message passing
\begin{equation}
\scalemath{0.9}{
    (\psi_{out,ij}^{(l_g)}, \phi_{p,ij}^{(l_g+1)}, \psi_{in,ij}^{(l_g)}) = g_1(\phi_{out,ij}^{(l_g)}, \phi_{p,ij}^{(l_g)}, \phi_{in,ij}^{(l_g)}).
    }
\end{equation}

\noindent Second, the aggregation step combines the information coming from all the edges of each node:
\begin{equation}
\scalemath{0.9}{
    \rho_i^{(l_g)} = \frac{1_g}{M_i} \Big(\sum_{j\in \mathcal{R}_{out}} \psi_{out,ij}^{(l_g)} + \sum_{j\in \mathcal{R}_{in}} \psi_{in,ji}^{(l_g)}\Big)
    }
\end{equation}
\noindent where $M_i$ is the number of edges for node $i$, and $\mathcal{R}_{out}$, $\mathcal{R}_{in}$ are the set of edges of the node as out(in)-bound objects. The resulting feature is fed to a final update MLP $g_2(\cdot)$
\begin{equation}
    \phi_i^{(l_g+1)} = \phi_i^{(l_g)} +  g_2(\rho_i^{(l_g)}).
\end{equation}

\subsection{Encoding a 3D Scene}
\label{sec:layout_en}

We respectively harness two parallel Graph Convolutional encoders $\mathcal{E}_\mathrm{layout}$,  and $\mathcal{E}_\mathrm{shape}$, for layout and shapes.
The layout encoder $\mathcal{E}_\mathrm{layout}$ is a GCN that takes the \emph{extended} graph $G_b$, where nodes $n_i=(o_i, b_i)$ are enriched with the set of 3D boxes $b$ for each object, and generates an output feature $f_{b,i}$ for each node $n_i$ with $f_{b} = \mathcal{E}_{layout}(G_b)$.

Though it is possible to sample shapes independently from the scene graph, it can lead to inconsistent configurations. For instance, we would expect an office chair to co-occur with a desk. As a consequence, we propose to leverage another GCN to infer consistent scene setups. While a loss directly on the bounding boxes works well, similarly learning a GCN Auto-Encoder on shapes, \eg point clouds, is a much more difficult task due to its uncontinuous output space. To circumvent this issue, we thus propose to instead learn how to generate shapes using a latent canonical shape space. This canonical shape space can be realized by various generative models having an encoder $\mathcal{E}_\mathrm{gen}(\cdot)$ and decoder $\mathcal{D}_\mathrm{gen}(\cdot)$, \eg by means of training an Auto-Encoder/Decoder~\cite{groueix2018atlas,park2019deepsdf}.
We create the extended graph $G_s$ with nodes $n_i = (o_i, e^s_i)$, where $e^s_i = \mathcal{E}_\mathrm{gen}(s_i)$. This formulation makes Graph-to-3D agnostic to the chosen shape representation. In our experiments, we demonstrate results with AtlasNet~\cite{groueix2018atlas} and DeepSDF~\cite{park2019deepsdf} as generative models. Please refer to the supplement for more details on AtlasNet and DeepSDF. Also here, we employ a GCN as shape encoder $\mathcal{E}_\mathrm{shape}$, which we feed with $G_s$ to obtain per node shape features $f_{s} = \mathcal{E}_{shape}(G_s)$.

\subsection{Shape and Layout Communication}\label{sec:shared}
As layout and shape prediction are related tasks, we want to encourage communication between both branches. Therefore, we introduce a shared encoder $\mathcal{E}_\mathrm{shared}$, which takes the concatenated output features of each encoder and computes a shared feature $f_\mathrm{shared} = \mathcal{E}_\mathrm{shared}(f_{bs}, \mathcal{R})$ with $f_{bs} = \{f_{b,i}\oplus f_{s,i} \text{ }| \text{ } i \in (1,...,N)\}$. Further, we feed the shared features to an MLP network to compute the shared posterior distribution $(\mu, \sigma)$ under a Gaussian prior. We sample $z_{i}$ from this distribution and feed the result to the associated layout and shape decoders. Since sampling is not differentiable, we apply the commonly used re-parameterization trick at training time to obtain $z_{i}$.

\subsection{Decoding the 3D Scene}
The layout decoder $\mathcal{D}_\mathrm{layout}$ is again a GCN having the same structure as the encoders. The last GCN layer is followed by two MLP branches, which predict box extents and location $b_{\text{-}\alpha,i}$ separately from angle $\alpha_i$. $\mathcal{D}_\mathrm{layout}$ is fed with a set of sampled latent vectors $z$, one for each node, within the learned distribution as well as the semantic scene graph $G$. It then generates the corresponding object 3D boxes $(\hat{b}_{\text{-}\alpha},\hat{\alpha}) = \mathcal{D}_\mathrm{layout}(z, \mathcal{O}, \mathcal{R})$.
The shape decoder $\mathcal{D}_\mathrm{shape}$ follows a similar structure as $\mathcal{D}_\mathrm{layout}$, with the difference that the GCN is followed by a single MLP producing the final shape encodings $\hat{e}^s = \mathcal{D}_\mathrm{shape}(z, \mathcal{O}, \mathcal{R})$. 

To obtain the final 3D scene, each object shape encoding is decoded into the respective shape $\hat{s}_i = \mathcal{D}_{gen}(\hat{e}^s_i)$. Each shape $\hat{s}_i$ is then transformed from canonical pose to scene coordinates, using the obtained bounding box $\hat{b}_i$.

\subsection{Scene Graph Interaction}
\label{sec:interaction}
\begin{figure}[t]
\begin{center}
   \includegraphics[width=0.88\linewidth]{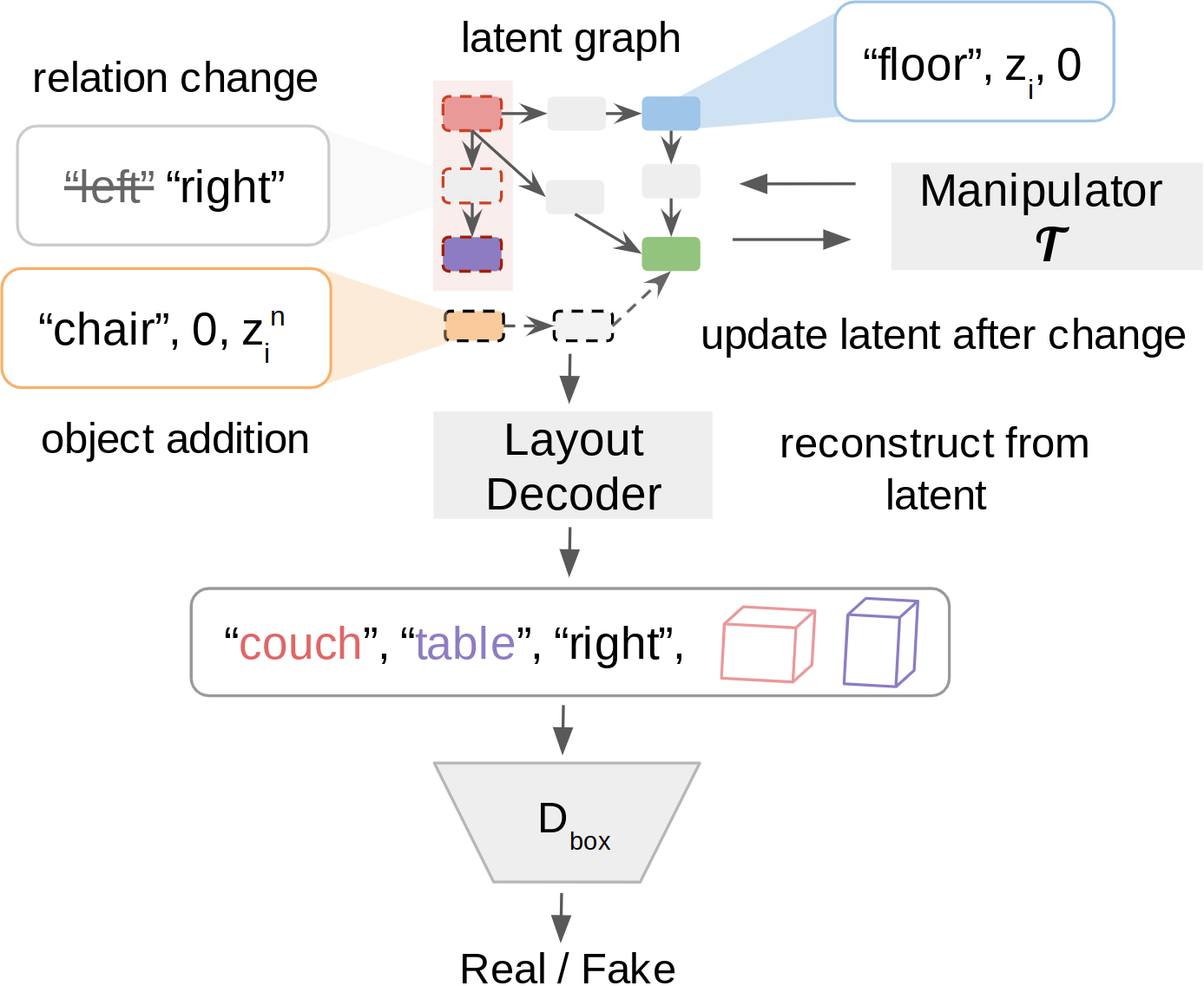}
\end{center}
   \caption{\textbf{Modifying scene graphs.} Given a scene graph we make changes in the nodes (object addition) or edges (relation change). Network $\mathcal{T}$ updates the latent graph accordingly. All edges that contain a change are passed to a relationship discriminator to encourage box prediction constrained on the node and edge labels.}
\label{fig:method_mani}
\end{figure}

To enable scene manipulation that is aware of the current scene, we extend our model with another GCN $\mathcal{T}$, directly operating on the shared latent graph $G_l=(z, \mathcal{O}, \mathcal{R})$ as obtained from the encoders.
First, we augment $G_l=(\hat{z},\widehat{\mathcal{O}}, \widehat{\mathcal{R}})$ with changes. Thereby, $\widehat{\mathcal{O}}$ is composed of the original nodes $\mathcal{O}$ together with the new nodes $\mathcal{O}'$ being added to the graph. Similarly, $\widehat{\mathcal{R}}$ consists of the original edges $\mathcal{R}$ together with the new out-going and in-going edges of $\mathcal{R}'$. Additionally, some edges of $\widehat{\mathcal{R}}$ are modified according to the input from the user. Finally, since we do not have any corresponding latent representations for $\mathcal{O}'$, we instead pad $z_i'$ with zeros to compute $\hat{z}_i$. Note that there can be infinitely possible outputs reflecting a given change. To capture this continuous output space, we concatenate $\hat{z}_i$ with samples $z^n_i$ from a normal distribution having zero mean and unit standard deviation, if the node has been part of a manipulation, otherwise we concatenate zeros. Then, the $\mathcal{T}$ network gives a transformed latent as $z_{\mathcal{T}} = \mathcal{T}( \hat{z} \oplus \hat{z}^n, \widehat{\mathcal{O}}, \widehat{\mathcal{R}})$, as illustrated in Figure~\ref{fig:method_mani}. Afterwards, the predicted latents for the affected nodes are plugged back into the original latent scene graph $G_l$. %
Finally, we 
feed the changed latent graph to the respective decoders to generate the updated scene, according to the changed scene graph. During inference, a user can directly make changes in the nodes and edges of a graph. At training time, we simulate the user input by creating a copy of the real graph exhibiting random augmentations, such as node addition, relationship label corruption, or alternatively, leave the scene unchanged. %

\subsection{Training Objectives}

The loss for training Graph-to-3D on the unchanged nodes, \ie generative mode and unchanged parts during manipulation, is composed of a reconstruction term
\begin{equation}
\scalemath{0.93}{
\begin{split}
\label{eq:loss_rec}
    \mathcal{L}_\mathrm{r} = \frac{1}{N}\sum_{i=1}^{N} (|| \hat{b}_{\text{-}\alpha,i} - b_{\text{-}\alpha,i}||_1 +  CE(\hat{\alpha}_i, \alpha_i) +
    ||\hat{e}^s_i - e^s_i||_1)
    \end{split}}
\end{equation}
\noindent and a Kullback-Leibler divergence term
\begin{equation}
\label{eq:kl}
\mathcal{L}_{KL} = D_{KL}(\mathcal{E}(z | G, \mathcal{B}, e^s) | p(z | G)),
\end{equation}
with $p(\cdot)$ denoting the Gaussian prior distribution and $\mathcal{E(\cdot)}$ being the complete encoding network. CE represents cross-entropy used to classify the angles, discretized in 24 classes. %
 
\subsubsection{Self-supervised Learning for Modifications}
\label{sec:discrim}
To train Graph-to-3D with changes, one requires appropriate pairs of scenes, \ie before and after interaction. Unfortunately, recording such data is very expensive and time consuming.
Furthermore, directly supervising the changed nodes with an $L_1$ loss is not an appropriate modeling for the one-to-many mapping of each relationship. Therefore, we propose the use of a novel relationship discriminator $D_{box}$, which can directly learn to interpret relationships and layouts from data, and ensure that the occasional relationship changes or node additions are correctly reflected in the 3D scene. %
We feed $D_{box}$ with two boxes, class labels, and their relationship. $D_{box}$ is then trained to enforce that the generated box will be following the semantic constraints from the relationship. To this end, we feed the discriminator with either real compositions or generated (fake) compositions, \ie boxes after modification. $D_{box}$ is then optimized such that it learns to distinguish between real and fake setups, whereas the generator tries to fool the discriminator by producing correct compositions under manipulations. The loss follows~\cite{NIPS2014_5423} and optimizes the following GAN objective
\begin{equation}
\scalemath{0.87}{
  \begin{split}
        \mathcal{L}_\mathrm{D,b} = & \min_G \max_D [ \sum_{(i,j)\in\mathcal{R}'}\mathbb{E}_{o_i, o_j, r_{ij},b_i, b_j}[log D_{box}(o_i, o_j, r_{ij},b_i, b_j)] \\ & + 
\mathbb{E}_{o_i, o_j, r_{ij}}[log(1-D_{box}(o_i, o_j, r_{ij}, \hat{b}_i, \hat{b}_j))] ].
\end{split}
}
\end{equation} 

 Notice that this discriminator loss is applied to all edges that contain a change.

With a similar motivation, we adopt an auxiliary discriminator~\cite{odena2017conditional} for the changed shapes, which in addition to the GAN loss, leverages a classification loss $\mathcal{L}_{aux}$ according to
\begin{equation}
\begin{split}
    \mathcal{L}_\mathrm{D,s} = \mathcal{L}_\mathrm{aux} + & \min_G \max_D [ \sum_{i=1}^{N} \mathbb{E}_{o_i,e^s_i}[log D_{shape}(e^s_i)] + \\&
\mathbb{E}_{o_i}[log(1-D_{shape}( \hat{e}^s_i))] ].
\end{split}
\end{equation}
\noindent Thereby, in addition to the real/fake decision, $D_{shape}$ predicts the class of the given latent shape encoding to encourage that the generated objects represent their true class, \ie $\mathcal{L}_\mathrm{aux}$ leverages the cross-entropy loss between the true $o_i$ class and the predicted class from $D_{shape}$. Therefore, the discriminator can learn the boundary of the underlying shape distribution and ensure that the reconstructed shape stems from this distribution.

\begin{table*}[t!]
    \centering
    \scalebox{0.75}{
    \begin{tabular}{l|c|ccccc|c}
    \midrule \midrule
        \multirow{2}{*}{Method} & Shape & left / & front / & smaller / & lower / & \multirow{2}{*}{same} & \multirow{2}{*}{total}  \\
        & Representation & right & behind & larger & higher & & \\
    \midrule \midrule
        3D-SLN \cite{Luo_2020_CVPR} & -- & 0.74 & 0.69 & 0.77 & 0.85 & \textbf{1.00} & 0.81 \\
        Progressive & -- & 0.75 & 0.66 & 0.74 & 0.83 &0.98 & 0.79 \\
        Graph-to-Box & -- & 0.82 & 0.78 & 0.90 & 0.95 & \textbf{1.00} & 0.89 \\
        Graph-to-3D & AtlasNet~\cite{groueix2018atlas} & \textbf{0.85} & 0.79 & 0.96 & 0.96 & \textbf{1.00} & 0.91 \\
        Graph-to-3D & DeepSDF~\cite{park2019deepsdf} & 0.81 & \textbf{0.81} & \textbf{0.99} & \textbf{0.98} & \textbf{1.00} & \textbf{0.92} \\
    \midrule \midrule
    \end{tabular}
    }
    \caption{Scene graph constrains on the \textbf{generation} task (higher is better). The total accuracy is computed as mean over the individual edge class accuracy to minimize class imbalance bias.}
    \label{tab:generation}
\end{table*}

\begin{table*}[t!]
    \centering
    \scalebox{0.75}{
    \begin{tabular}{l|c|c|ccccc|c}
    \midrule \midrule
        \multirow{2}{*}{Method} & Shape  & \multirow{2}{*}{mode}  & left / & front / & smaller / & lower / & \multirow{2}{*}{same} & \multirow{2}{*}{total}  \\
        & Representation & & right & behind & larger & higher & & \\
    \midrule \midrule
        3D-SLN \cite{Luo_2020_CVPR} & \multirow{3}{*}{--} & \multirow{8}{*}{change} & 0.62 & 0.62 & 0.66 & 0.67 &0.99 & 0.71  \\
        Progressive & & & \textbf{0.81} & \textbf{0.77} & 0.76 & \textbf{0.84} & \textbf{1.00} & \textbf{0.84} \\
        Graph-to-Box & & & 0.65 & 0.66 & 0.73 & 0.74 &0.98 & 0.75 \\
        \cmidrule{1-2}
        Graph-to-3D w/o $\mathcal{T}$ & \multirow{2}{*}{AtlasNet~\cite{groueix2018atlas}} & & 0.64 & 0.66 & 0.71 & 0.78 & 0.96 & 0.75 \\
        Graph-to-3D & & &  0.73 & 0.67 & \textbf{0.82} & 0.79 & \textbf{1.00} & 0.80 \\
        \cmidrule{1-2}
        Graph-to-3D w/o $\mathcal{T}$ & \multirow{2}{*}{DeepSDF~\cite{park2019deepsdf}} & &  0.71 & 0.71 & 0.80 & 0.79 & 0.99 & 0.80 \\
        Graph-to-3D &  & & 0.73 & 0.71 & \textbf{0.82} & 0.79 & \textbf{1.00} & 0.81 \\
        \midrule \midrule
        3D-SLN \cite{Luo_2020_CVPR} & \multirow{3}{*}{--} & \multirow{8}{*}{addition} & 0.62 & 0.63 & 0.78 & 0.76 & 0.91  & 0.74 \\
        Progressive & &  &  \textbf{0.91} & \textbf{0.88} & 0.79 & \textbf{0.96} & \textbf{1.00} & \textbf{0.91} \\
        Graph-to-Box &  & & 0.63 & 0.61 & 0.93 & 0.80 & 0.86  &  0.76 \\
        \cmidrule{1-2}
        Graph-to-3D w/o $\mathcal{T}$ & \multirow{2}{*}{AtlasNet~\cite{groueix2018atlas}} & & 0.64 & 0.62 & 0.85 & 0.84 & \textbf{1.00} & 0.79 \\
        Graph-to-3D &  & & 0.65 & 0.71 & 0.96 & 0.89 & \textbf{1.00} & 0.84 \\
        \cmidrule{1-2}
        Graph-to-3D w/o $\mathcal{T}$ & \multirow{2}{*}{DeepSDF~\cite{park2019deepsdf}} & & 0.70 & 0.73 & 0.85 & 0.88 &0.97 & 0.82 \\
        Graph-to-3D &  & & 0.69 & 0.73 & \textbf{1.00} & 0.91 &0.97 & 0.86\\
    \midrule \midrule
    \end{tabular}}
    \caption{Scene graph constraints on the \textbf{manipulation} task (higher is better). The total accuracy is computed as mean over the individual edge class accuracy to minimize class imbalance bias. Top: Relationship change mode. Bottom: Node addition mode.}
    \label{tab:mani}
\end{table*}

To summarize, our final loss becomes
\begin{align}
    \mathcal{L}_\mathrm{total} = \mathcal{L}_\mathrm{r} + \lambda_{KL} \mathcal{L}_\mathrm{KL} + \lambda_{D,b} \mathcal{L}_\mathrm{D,b} + \lambda_{D,s} \mathcal{L}_\mathrm{D,s}
\end{align}
\noindent where the $\lambda$s refer to the respective loss weights. We refer to the supplementary material for implementation details.

\section{Results}
\begin{table*}[t!]
    \centering
    \scalebox{0.75}{
    \begin{tabular}{lc|c|cccc|cccc}
        \midrule \midrule
         \multirow{2}{*}{Method} &  \multirow{2}{*}{Shape Model} & Shape & \multicolumn{4}{c|}{Generation} & \multicolumn{4}{c}{Manipulation} \\
        & & Representation & Size & Location & Angle & Shape & Size & Location & Angle & Shape \\
        \midrule  \midrule
        3D-SLN~\cite{Luo_2020_CVPR} & Retrieval & \multirow{2}{*}{3RScan Data} & 0.026 & 0.064 & 11.833	& \textbf{0.088} & 0.001 & 0.002 & 0.290 & 0.002 \\
        Progressive & -- & & 0.009 & 0.011  & 1.494 & -- & 0.008 & 0.008 &  1.559 & --  \\
        \midrule
        Graph-to-Box & Graph-to-Shape & \multirow{2}{*}{AtlasNet~\cite{groueix2018atlas}} & 0.009 & 0.024 & 1.869 & 0.000 &  0.007 &0.019 &  2.920 & 0.000  \\
        \multicolumn{2}{c|}{Graph-to-3D}  & & \textbf{0.097} & \textbf{0.497}  & \textbf{20.532} &  0.005 & \textbf{0.037} & \textbf{0.061} & \textbf{14.177} & 0.007\\
        \midrule
        Graph-to-Box & Graph-to-Shape & \multirow{2}{*}{DeepSDF~\cite{park2019deepsdf}} & 0.009 & 0.024 & 1.895 & 0.011 & 0.005 & 0.019 & 3.391 & 0.014 \\
        \multicolumn{2}{c|}{Graph-to-3D} & & 0.091 & 0.485 & 19.203 & 0.015 & 0.015 & 0.035 & 9.364 & \textbf{0.016} \\
        \midrule \midrule
    \end{tabular}}
    \caption{Comparison on diversity results (std) on the generation (left) and manipulation tasks (right), computed as standard deviation over location and size in meters and angles in degrees. For shape we report the average chamfer distance between consecutive generations.}
    \label{tab:diversity}
\end{table*}

\begin{figure*}[t]
\begin{center}
   \includegraphics[width=0.95\linewidth]{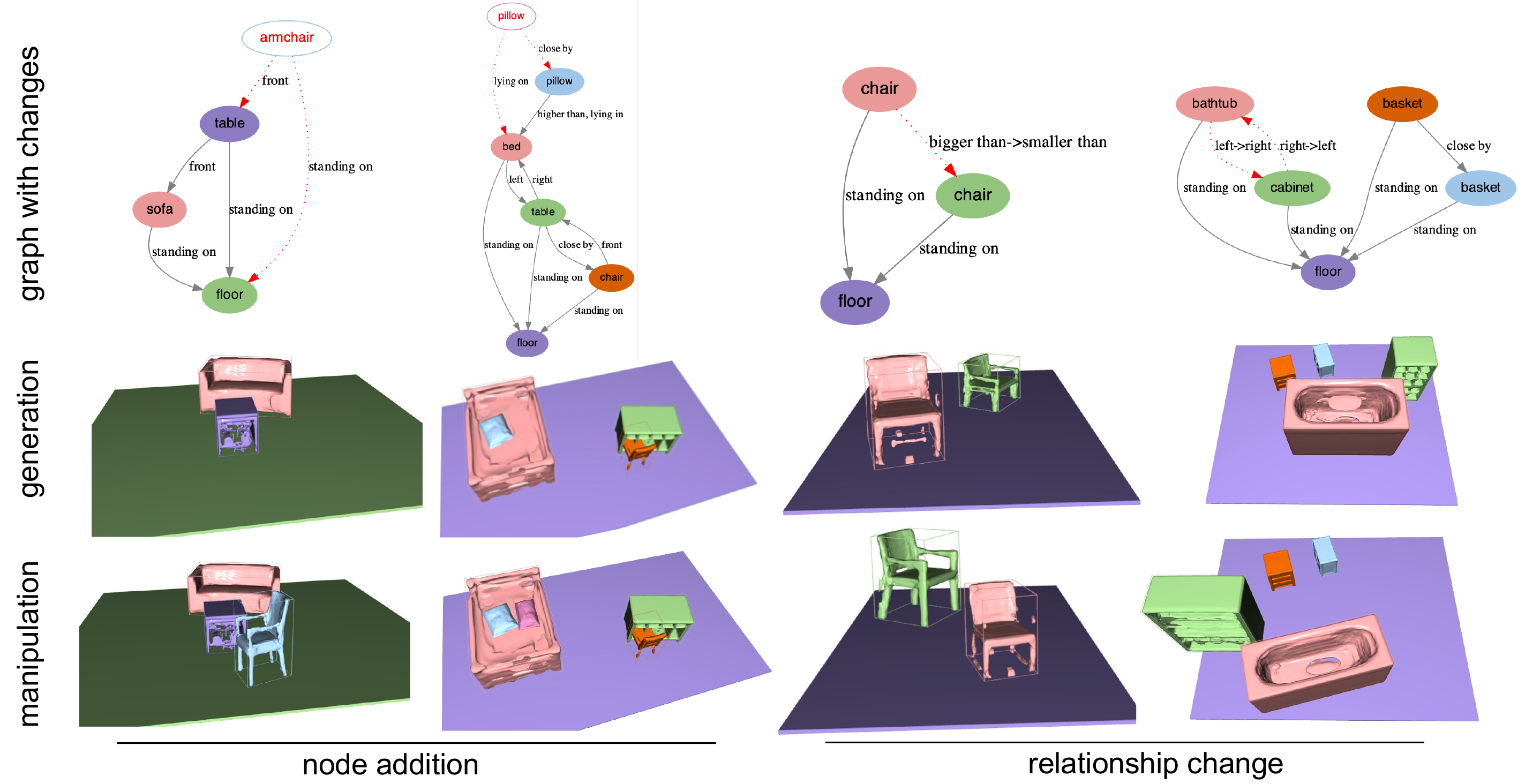}
\end{center}
   \caption{\textbf{Qualitative results} of Graph-to-3D (DeepSDF encoding) on 3D scene generation (middle) and manipulation (bottom), starting from a scene graph (top). Dashed lines reflect new/changed relationship, while empty nodes indicate added objects.
   }
\label{fig:results}
\end{figure*}

\begin{figure*}[t]
\begin{center}
   \includegraphics[width=0.9\linewidth]{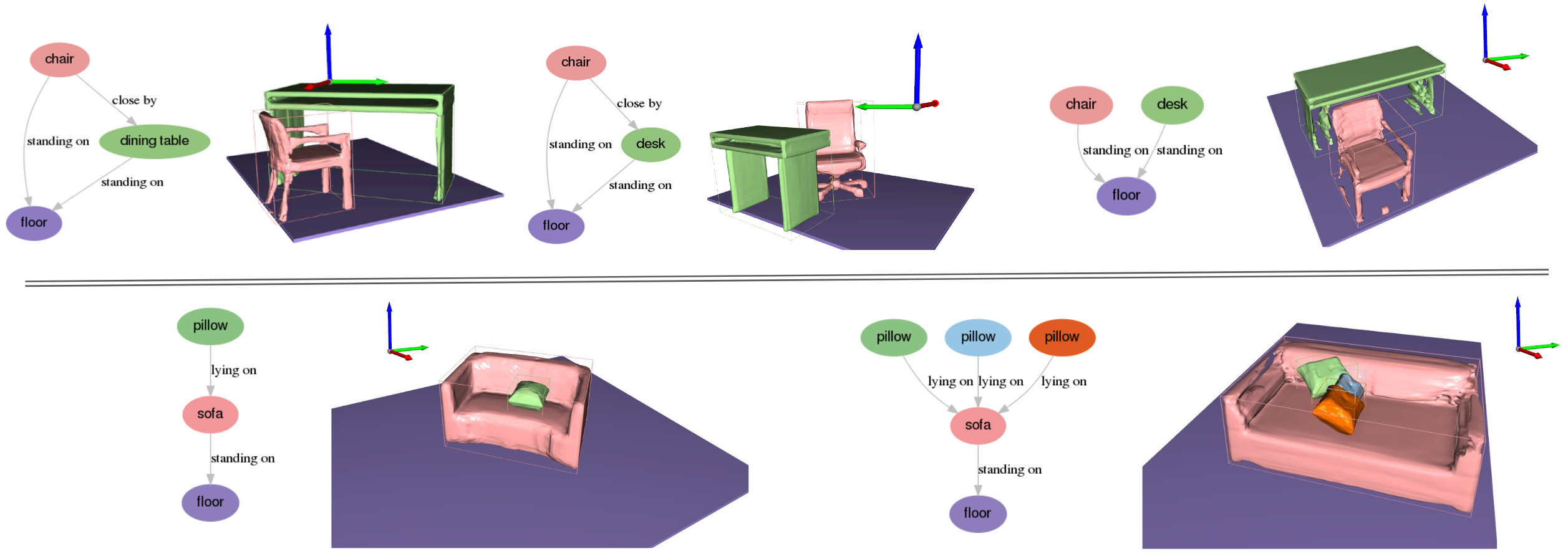}
\end{center}
   \caption{\textbf{Effect of scene context in scene generation.} \emph{Top:} Connection to a desk makes a chair look like an office chair. \emph{Bottom:} The number of pillows lying on a sofa affects its size and style. 
   }\label{fig:results_context}
\end{figure*}

In this section we describe the evaluation we used to assess the performance of the proposed approach in terms of plausible layout and shape generation that meets the constraints imposed by the input scene graph. %

\subsection{Evaluation protocol}
We evaluate our method on the official splits of 3DSSG dataset~\cite{3DSSG2020}, with 160 object classes and 26 relationship classes. Since we expect multiple possible results for the same input, typical metrics, such as $L_1$/$L_2$ norm or Chamfer loss are not suitable, due to the strict comparison between the predictions and the ground truth. Following~\cite{Luo_2020_CVPR} we rely on geometric constraints to measure if the input relationships are correctly reflected in the generated layouts. We test the constraint metric on each pair of the predicted boxes that are connected with the following relationships: left, right, front, behind, smaller, larger, lower, higher and same (\cf supplementary material for more details). 

As a way to quantitatively evaluate the generated scenes and shapes, we perform a cycle-consistency experiment. Given the generated shapes from our models, we predict the scene graph, using the state-of-the-art scene graph prediction network (SGPN) from~\cite{3DSSG2020}. We then compare the ground truth scene graphs (\ie input to our models) against the predicted graphs from SGPN. We base this comparison on the standard top-k recall metric for objects, predicates and relationship triplets from~\cite{3DSSG2020} (see supplement). This is motivated by the expectation that plausible scenes should result to the same graph as the input graph. Similar metrics have been utilized for image generation from semantics~\cite{wang2018pix2pixHD}, using the inferred semantics from the generated image. In addition, in the supplement we report a user study to assess the global correctness and style fitness.%

\subsection{Baselines}

\noindent \textbf{3D-SLN} %
With the unavailability of SunCG, we train \cite{Luo_2020_CVPR} on 3DSSG using their official code repository. As we do not focus on images, we omit the rendering component. To obtain shapes for 3D-SLN, we follow their retrieval approach, in which for every $\hat{b}_i$ we retrieve from 3RScan the object shape from the same class, with the highest similarity.

\noindent \textbf{Progressive Generation}
A model which naturally supports 3D generation and manipulation would be a progressive (auto-regressive) approach, as also explored in \cite{Wang2019PlanITPA} for room planning. At each step a GCN (same as $\mathcal{D}_{layout}$) receives the current scene, together with a new node $n_a$ to be added. We refer the reader to the supplement for more details on the progressive baseline. %

\noindent \textbf{Ablations} To ablate the relevance of using a GCN for the shape generation, we leverage a variational autoencoder directly based on AtlasNet, without awareness of the neighbouring objects. We provide more details in the supplement.
Further, we ablate the sharing of layout and shape, by training a model with separate GCN-VAEs for shape (Graph-to-Shape) and layout (Graph-to-Box), which follow the same architecture choices, except $\mathcal{E}_\mathrm{shared}$. We also run our method without modification network $\mathcal{T}$.

\begin{table*}[t!]
    \centering
    \scalebox{0.75}{
        \begin{tabular}{ll|c|ccc|ccc|ccc}
            \midrule \midrule
            \multirow{2}{*}{Layout Model} & \multirow{2}{*}{Shape Model} & Shape & \multicolumn{3}{c|}{Recall Objects} &\multicolumn{3}{c|}{Recall Predicate} & \multicolumn{3}{c}{Recall Triplets} \\ 
            & & Representation & Top 1 & Top 5 & Top 10 & Top 1 & Top 3 & Top 5 & Top 1 & Top 50 & Top 100 \\
            \midrule \midrule
            3D-SLN \cite{Luo_2020_CVPR} & Retrieval & \multirow{2}{*}{3RScan Data} & \textbf{0.56} & 0.81 & 0.88 & 0.50 & \textbf{0.82} & 0.86 & 0.15 & 0.57 & 0.82 \\
            Progressive & Retrieval & & 0.35 & 0.66 & 0.79 &  0.41 &  0.70 & 0.82 & 0.09 &  0.40 & 0.70 \\
            \midrule
            Graph-to-Box & AtlasNet VAE & \multirow{4}{*}{AtlasNet~\cite{groueix2018atlas}} & 0.41 & 0.74 & 0.83 & 0.57 & 0.80 & 0.88 & 0.08 & 0.46 & 0.77 \\
            $^{\ddag}$Graph-to-Box &  $^{\ddag}$Graph-to-Shape & & 0.39 & 0.68 & 0.77 & 0.55 &  0.79 & 0.88 & 0.05 & 0.35 & 0.69 \\
            Graph-to-Box & Graph-to-Shape &  & 0.51 & 0.81 & 0.86 & 0.57 & 0.80 & 0.88 & \textbf{0.23} & 0.63 & 0.84 \\
            \multicolumn{2}{c|}{Graph-to-3D} &  & 0.54 & \textbf{0.84} & \textbf{0.90}  & \textbf{0.60} & \textbf{0.82} & \textbf{0.90} & 0.21 & \textbf{0.65} & \textbf{0.85} \\
            \midrule
            Graph-to-Box & Graph-to-Shape &\multirow{2}{*}{DeepSDF~\cite{park2019deepsdf}} & 0.47 & 0.74 & 0.83 & 0.57 & 0.80 & 0.87 & 0.14 & 0.57 & 0.81 \\
            \multicolumn{2}{c|}{Graph-to-3D}  &  & 0.51 & 0.80 & 0.88  & 0.58 & 0.80 & 0.89 & 0.19 & 0.59 & 0.83 \\
            \midrule
             \multicolumn{3}{c|}{3RScan data}  & 0.53 & 0.82 & 0.90 & 0.75 & 0.93 & 0.98 & 0.18 & 0.61 & 0.83
             \\
            \midrule
            \midrule

        \end{tabular}
    }
    \caption{Scene graph prediction accuracy on 3DSSG, using the SGPN model from \cite{3DSSG2020}, measured as top-k recall for object, predicate and triplet prediction (higher is better). $^{\ddag}$Model trained with non-canonical objects, exhibiting significantly worse results.}
    \label{tab:sup_s2g_large}
\end{table*}

\subsection{Layout evaluation}
Table \ref{tab:generation} reports the constrain accuracy metric on the generative task. We observe that Graph-to-3D outperforms the baselines as well as the variant decoupled layout and shape Graph-to-box on all metrics.
Table~\ref{tab:mani} evaluates the constraint accuracy metric on the manipulation task. We report the node addition experiment and the relationship change experiment separately. We observe that the progressive model performs best for node addition (Table~\ref{tab:mani}, bottom), while ours is fairly comparable for changes. This is natural as the progressive model is explicitly trained for addition. %
The models using $\mathcal{T}$ perform better than 3D-SLN or the respective model without $\mathcal{T}$ on the manipulation task, which is expected since these approaches explicitly model an architecture that supports such changes.

In addition, we measure diversity as standard deviation among 10 samples that are generated under the same input. We compute this metric separately over each bounding box parameter, and compute the mean over size, translation in meters and angle in degrees. To measure shape diversity, we report the average chamfer distance between these 10 samples. Results are shown in Table \ref{tab:diversity}. The progressive generation shows the lowest values in diversity for both generation and modification. The other models, on the other hand, exhibit more interpretable diversity results, with larger values for position than for object size. Nevertheless, both shared models come out superior for diversity in layout. As for shape, the two shared models are again superior for manipulation, yet, we perform a bit worse for generation.

\subsection{Shape evaluation}
Figure~\ref{fig:results} shows qualitative results from Graph-to-3D. We first sample a scene conditioned on a scene graph (top), and then apply a change in the graph which is then reflected in the scene. The model understands diverse relationships such as support (lying on), proximity (left, front) and comparison (bigger than). For instance, the model is able to place a pillow on the bed, or change chair sizes in accordance with the edge label. In addition, the object shapes and sizes well represent the class categories in the input graph.

In Figure~\ref{fig:results_context} we illustrate the effect of scene context on shape generation. For instance, chairs tend to have an office style (middle) while connected to a desk, and a more standard style when connected to a dining table (left), or when there is no explicit connection to the desk (right). In addition, having many pillows on a sofa contributes to its style and larger size. These patterns learned from data show another interesting advantage of the proposed graph-driven approach based on learned shapes.   

The quantitative results on 3D shapes and complete 3D scenes are shown on Table~\ref{tab:sup_s2g_large}. The object and predicate recall metric is mostly related to namely shape generation and layout generation quality. The triplet recall measures the combined influence of all components. The table compares different shape models, such as AtlasNet VAE, Graph-to-Box/Shape and our shared model Graph-to-3D. For reference we present the scene graph prediction results on the ground truth scenes (3RScan data). As expected, the latter has the highest accuracy in predicate prediction. Interestingly, on metrics that rely on shapes, it is comparable to our Graph-to-3D model. %
Models based on a GCN for shape generation outperform the simple AtlasNet VAE, that does not consider inter-object relationships. Comparing the shared and disentangled models we observe that there is a consistent performance gain for both the layout generation as well as shape, meaning that the two tasks benefit from the joint layout and shape learning. Finally, we also run our baseline Graph-to-Box/Shape using shapes in non-canonical pose. The performance of this model drops significantly, demonstrating the relevance of our annotations. %

\section{Conclusion}

In this work, we propose Graph-to-3D a novel model for end-to-end 3D scene generation and interaction using scene graphs, and explored the advantages of joint learning of shape and layout. We show that the same model can be trained with different shape representations, including point clouds and implicit functions (SDFs). %
Our evaluations on quality, semantic constrains and diversity show compelling results on both tasks. Future work will be dedicated to generating objects textures, combined with scene graph attributes that describe visual properties. %

\section{Acknowledgements}

This research work was supported by the Deutsche Forschungsgemeinschaft (DFG), project 381855581. We thank all the participants of the user study.

{\small
\bibliographystyle{ieee_fullname}
\bibliography{egbib}
}

\clearpage
\setcounter{section}{0}
\renewcommand*{\thesection}{\Alph{section}}
\section{Supplementary material}
This document supplements our main paper entitled \emph{Graph-to-3D: End-to-end Generation and Manipulation of 3D Scenes using Scene Graphs} by providing i) more details on the data preparation and ii) inference mode. We iii) give additional information on the employed GCN and discriminators, as well as iv) shape generation networks. Further, we provide v) more details on the employed baselines. We vi) clarify the used metrics, \ie define the computation of our scene graph constraints used in the layout evaluation and provide details of the top-K recall. Finally, vii) we report the results of a user study and viii) demonstrate qualitative and diversity results for the generated 3D scenes.

\subsection{Data preparation and annotation}

\begin{figure*}[t]
\begin{center}

   \includegraphics[width=\linewidth]{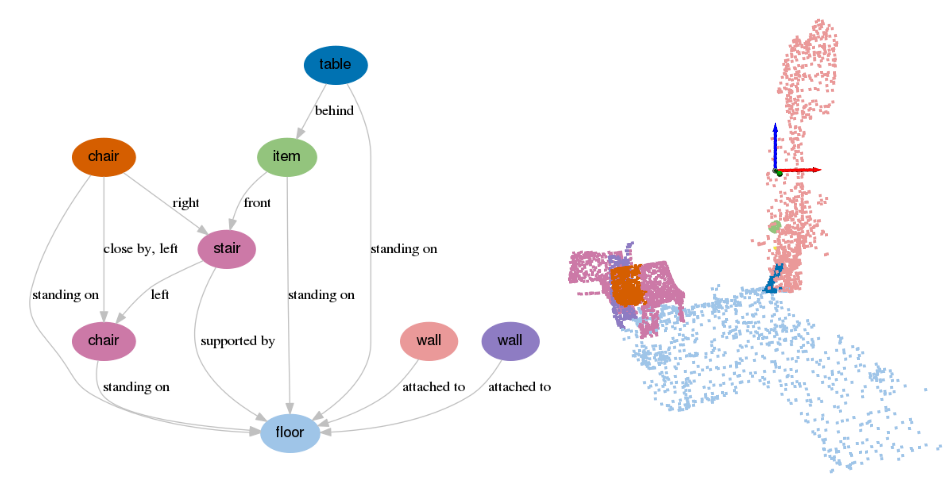}
\end{center}
   \caption{Example of ground truth graph from 3DSSG and the respective 3D scan from the 3RScan dataset.}
\label{fig:sup_gt_data}
\end{figure*}

In this section we provide more details on our data preparation pipeline, used to obtain and refine oriented bounding boxes for 3RScan objects (\cf Figure \ref{fig:sup_gt_data} for a common ground truth scene example that illustrates the reconstruction partiality, as well as the respective scene graph from 3DSSG).

\paragraph{Extraction of 3D bounding boxes}

\begin{figure*}[t]
\begin{center}

   \includegraphics[width=\linewidth]{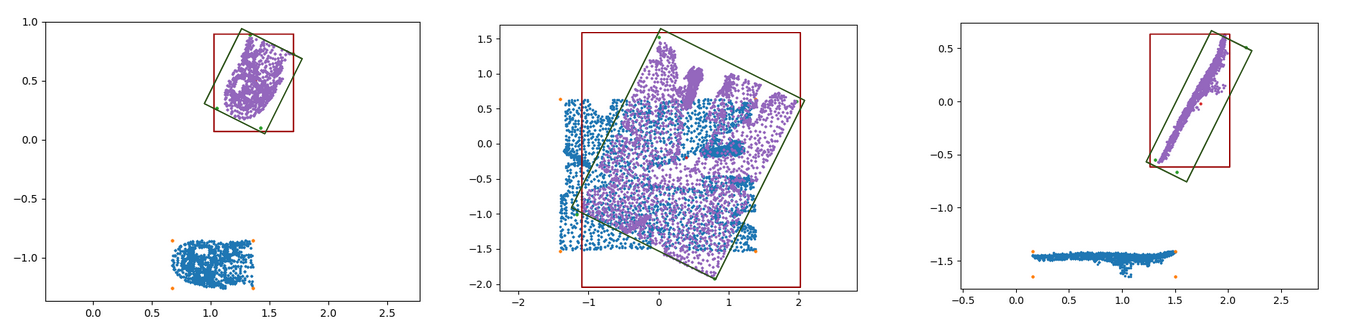}
\end{center}
   \caption{\textbf{Data preparation} (Top-down view of 3D point clouds). Violet: Original point cloud rotation. Blue: Point cloud in rotation that gives the smallest axis-aligned surface area. Red: axis-aligned box. Green: Oriented bounding box resulting from our data preparation. }
\label{fig:sup_data}
\end{figure*}

Figure \ref{fig:sup_data} illustrates the oriented bounding box preparation pipeline presented in the main paper, in top-down view. Given the original point cloud of an object (violet), the algorithm identifies the point cloud rotation (blue) that leads to smallest surface area for the axis-aligned bounding box. This rotation is then used to transform the identified box back in the original point cloud coordinates (green).  

\paragraph{Canonical pose annotation} We map the 160 object classes in 3DSSG~\cite{3DSSG2020} to the RIO27 label set from 3RScan~\cite{Wald2019RIO} and divide them in three categories, based on their symmetry properties. Table~\ref{tab:annotation_cat} gives the full object class list for each annotation category.
\begin{enumerate}
    \item \emph{Objects with two symmetry axes} such as tables, bathtubs, desks, walls, are annotated automatically, considering the direction with the largest size as front. 
    \item For a subset of objects, such as cabinet, shelf and oven, we annotate automatically based on the following observation. Given that such objects are usually attached to a vertical surface (wall) the 3D reconstruction for their back side is missing. Therefore, we first apply the rule of subset 1. to identify the directionless axis and then define the front side of the object as the direction where the center of mass is leaning towards. 
    \item \emph{Objects with one symmetry axis} such as chair, sofa, sink, bed are annotated manually. The annotator is presented with the object point cloud, inside an oriented bounding box, and is given four choices regarding the front direction of the object. 

\end{enumerate}

\begin{table}[h!]
    \centering
    \begin{tabular}{c|l}
    \toprule
Category & Classes \\ \midrule
\multirow{2}{*}{1} & table desk wall floor door window  tv\\        
  & curtain ceiling box bathtub object \\ \midrule
\multirow{2}{*}{2} & cabinet nightstand shelf fridge lamp \\
  & blanket clothes oven towel pillow \\ \midrule
3 & chair sofa bed toilet sink \\ \bottomrule
    \end{tabular}
    \caption{Annotation categories mapped to RIO27 label set.}
    \label{tab:annotation_cat}
\end{table}

\subsection{Inference}
\paragraph{Generation} Given a scene graph, we first sample a random vector per-node from the gaussian prior. Then we feed the augmented scene graph (class embeddings and sampled vectors) to the shape and layout decoders to recover a 3D scene. 

\paragraph{Manipulation} We first encode the input scene given the scene graph (newly added nodes are again sampled from the gaussian prior). We then run $\mathcal{T}$ to update the latent of the changed nodes \wrt the new graph, decode the scene and add the changes to the input scene.

\subsection{Implementation details}
We use 5 layers for each GCN model. In the encoders $\mathcal{E}_\mathrm{shape}$ and $\mathcal{E}_\mathrm{layout}$, prior to the GCN computation, the class categories $o_i$ and $r_{ij}$ are fed to embedding layers, while the shape embedding, box and angles are projected through a linear layer. All discriminators consist of fully-connected layers, where all layers apart from the last one are followed by batch norm and Leaky-ReLU. For $D_{box}$ (Table \ref{tab:d_box}), consisting of 3 layers, the last fully-connected layer is followed by a sigmoid. Here the class categories $o_i$ and $r_{ij}$ are fed in one-hot form giving a size of 160 and 26. For $D_{shape}$ (Table \ref{tab:d_shape}) after two consecutive layers, we employ two branches of fully-connected layers, followed by namely a softmax (for classification, outC) and sigmoid (for discrimination, outD).  
We use the Adam optimizer with a learning rate of $0.001$ and batch size of $8$, to train the model for 100 epochs. The training takes one day on one Titan Xp GPU. The loss weights are set to $\lambda_{KL}=0.1$, $\lambda_{D,b}=0.1$ and $\lambda_{D,s}=0.1$. The shape embeddings $e^s_i$ have a size of 128. 

\begin{table}[h]
    \centering
    \scalebox{0.9}{
    \begin{tabular}{c|c|c|c|c}
    \toprule
        layer & layer & input & input & output \\
          id & type & layer & channels & channels \\ \midrule
         L1 & Linear & $(o_i, o_j, r_{ij}, b_i, b_j)$ & 360 & 512 \\
         L2 & Batch Norm & L1 & 512  & 512 \\
         L3 & Leaky-ReLU & L2 & 512 &  512\\
         L4 & Linear  & L3 & 512 & 512 \\
         L5 & Batch Norm & L4 & 512  & 512 \\
         L6 & Leaky-ReLU & L5 & 512 &  512 \\
         L7 & Linear & L6 & 512 & 1 \\
         out & Sigmoid & L7 & 1 & 1 \\
         \bottomrule
    \end{tabular}}
    \caption{Architecture of $D_{box}$}
    \label{tab:d_box}
\end{table}

\begin{table}[h]
    \centering
    \scalebox{0.9}{
    \begin{tabular}{c|c|c|c|c}
    \toprule
        layer & layer & input & input & output \\
          id & type & layer & channels & channels \\ \midrule
         L1 & Linear & $e^s_i$ & 128 & 512 \\
         L2 & Batch Norm & L1 & 512  & 512 \\
         L3 & Leaky-ReLU & L2 & 512 &  512\\
         L4 & Linear  & L3 & 512 & 512 \\
         L5 & Batch Norm & L4 & 512  & 512 \\
         L6 & Leaky-ReLU & L5 & 512 &  512 \\ \midrule
         L7 & Linear & L6 & 512 & 1 \\
         outD & Sigmoid & L7 & 1 & 1 \\ \midrule
         L9 & Linear & L6 & 512 & 160 \\
         outC & Softmax & L9 & 160 & 160 \\
         \bottomrule
    \end{tabular}}
    \caption{Architecture of $D_{shape}$}
    \label{tab:d_shape}
\end{table}

\subsection{Generation of 3D scenes $\mathcal{F}_{gen}$}
\label{sec:pointgen}
 
\paragraph{Point clouds} We base our point cloud approach on AtlasNet~\cite{groueix2018atlas}. In particular, we employ AtlasNet to learn a low-dimensional latent space embedding on the point clouds. AtlasNet is grounded on PointNet~\cite{qi2016pointnet} and consumes a whole point cloud which it then encodes into a global feature descriptor $\mathcal{E}_{atlas}$. AtlasNet is particularly suited since the sampling on the uv-map allows to generate point clouds at arbitrarily resolutions while only using a small set of points during training. This significantly speeds up training while saving memory, thus allowing larger batch sizes. The 3D point cloud can be inferred by using this global feature descriptor together with sampled 2D points from the aforementioned uv-map running them through the decoder $\mathcal{D}_{atlas}$. We train AtlasNet on a mixture of synthetic data from ShapeNet and real 3RScan objects in canonical pose.%

\paragraph{Implicit functions}
In addition, we also employ implicit functions as shape representation using DeepSDF~\cite{park2019deepsdf}. To this end, we train an individual Auto-Decoder for each class using ShapeNet~\cite{chang2015shapenet}. Thereby, we use 350 shapes in canonical pose and learn a 128-dimensional continuous shape space. We then label each object in 3RScan with the best fitting descriptor. Initially, we attempted to use a similar partial scan alignment as originally proposed in~\cite{park2019deepsdf}. Yet, this did not work well in practice as the point quality was too low. Hence, we instead simply queried each learned descriptor from our shape space with the 3D points of the object, and labeled the object with the descriptor giving minimal average error in SDF. Notice that since we learn a generative model on top of these labels, Graph-to-3D can still exploit the full potential of the continuous shape space.

\subsection{Baseline details}
\paragraph{Variational AtlasNet}
For the variational AtlasNet (shape model without without GCN) we enforce a Gaussian distribution onto the embedding space of AtlasNet (AtlasNet VAE). In this model, the shapes are generated without awareness of the neighbouring objects. For a given point cloud $s_i$ we can compute the posterior distribution $(\mu_i, \sigma_i)=\mathcal{E}_{gen}(s_i)$ with $(\mu, \sigma)$ being the mean and log-variance of a diagonal Gaussian distribution. Sampling from the posterior allows to generate on the fly new shapes during inference. 

\paragraph{Progressive model} This baseline receives at each step the current scene, together with a new node $n_a$ to be added. Thereby, for the current scene nodes $n$, the model $\mathcal{A}$ receives the 3D boxes $b$ as well as the category labels $o$ for nodes and edges $r$ and predicts the new box according to $b_i = \mathcal{A}(o_i, r_ij, z_i, o, r, z)$. Here $z_i$ denotes randomly sampled noise from a normal distribution with zero mean and unit standard deviation. Note that for the new node $n_i$, we only feed the object category $o_i$ as well as its relationships $r_ij$ with existing objects $j$.
During inference, the method assumes the first node given, then gradually adds nodes and connections. In manipulation mode, the method receives a ground truth scene and a sequence of novel nodes to be added. We train the progressive baseline with varying graph sizes (2-10), such that it can learn to predict the consecutive node for different generation steps. We order the nodes based on the graph topology of the support relationships, \eg pillow is generated after the supporting bed. In addition, we place the disconnected nodes last in order. 

\subsection{Metrics}

\paragraph{Scene graph constraints}

For layout evaluation \emph{w.r.t} the employed scene graph constraints, our metrics follow the definitions from Table~\ref{tab:constraints}. Though ideally we want to validate all edges in 3DSSG, not all of them can be captured with a geometric rule, as they are manually annotated  (\eg \emph{belonging to}, \emph{leaning against}).

\begin{table}[h!]
    \centering
    \begin{tabular}{c|c}
    \toprule
    Relationship & Rule \\ \midrule
    left of    &  $c_{x,i} < c_{x,j}$ and $\mathrm{iou}(b_i,b_j)<0.5$ \\
    right of   & $c_{x,i} > c_{x,j}$ and $\mathrm{iou}(b_i,b_j)<0.5$ \\ \midrule
    front of & $c_{y,i} < c_{y,j}$ and $\mathrm{iou}(b_i,b_j)<0.5$\\
    behind of & $c_{y,i} > c_{y,j}$ and $\mathrm{iou}(b_i,b_j)<0.5$ \\ \midrule
    higher than &  $h_i + c_{z,i} / 2 > h_j + c_{z,j} / 2$\\
    lower than & $h_i + c_{z,i} / 2 < h_j + c_{z,j} / 2$\\ \midrule
    smaller than & $w_i l_i h_i < w_j l_j h_j$ \\
    bigger than & $w_i l_i h_i > w_j l_j h_j$ \\ \midrule
    same as & $\mathrm{iou}_C(b_i, b_j) > 0.5 $\\
    \bottomrule
    \end{tabular}
    \caption{Computation of geometric constraint accuracy, for two instances $i$ and $j$. $iou_C$ refers to iou computation after both objects have been 0-centered.}
    \label{tab:constraints}
\end{table}

\begin{figure*}[t!]
    \centering
    \includegraphics[width=\linewidth]{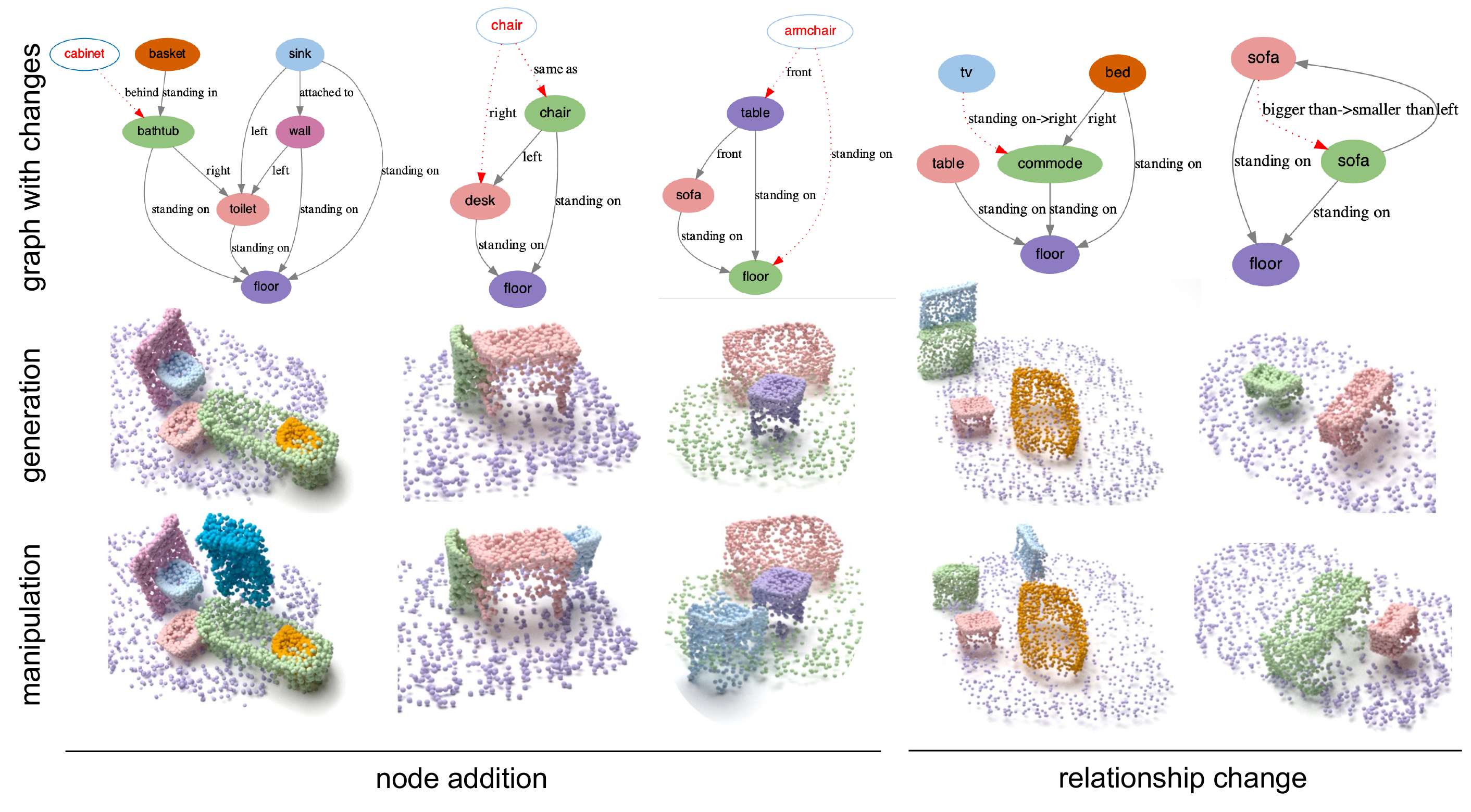} 
    \caption{Generation (middle) and manipulation (bottom) of full 3D scenes from scene graphs (top) for the Graph-to-3D model based on AtlasNet encodings for shape. The graph also contains the applied changes, in the form of dashed lines for new/changed relationship, and empty nodes for added objects.}
    \label{fig:pcd_comparison}
\end{figure*}

\paragraph{Top-K recall} We utilize the same top-K recall metric as in 3DSSG~\cite{3DSSG2020} to evaluate the SGPN predictions. For each object node or predicate, the top-K metric checks if the ground truth label is within the corresponding top k classification scores. To obtain a triplet score, we multiply the scores of the two respective objects as well as the relationship predicate. Then, similarly, we check if the ground truth triplet is among the top-K scores. 

\subsection{User study} We conducted a perceptual study with 20 people evaluating $\approx$30 scenes each. Each sample features a scene graph, the 3D-SLN~\cite{Luo_2020_CVPR} baseline with retrieved shapes from ShapeNet and our shared model (given anonymously and in random order). The users then rated each scene 1-7 on three aspects 1) global correctness, 2) functional and style fitness between objects and 3) correctness of graph constraints. 3D-SLN reported $2.8, 3.7, 3.6$, respectively, while ours exceeded them with $4.6, 4.9, 5.4$. Our method was preferred in namely $72\%, 62\%, \text{ and } 68\%$ of the cases.

\subsection{Additional Qualitative Results}

In this section we demonstrate additional qualitative results for 3D layouts and full 3D scenes with shapes from AtlasNet~\cite{groueix2018atlas} as well as DeepSDF~\cite{park2019deepsdf}. We would like to emphasize that in our manipulation experiments, we intentionally allow the network to also change the shape of the objects that are involved in a relationship change, to demonstrate diversity. Nonetheless, notice that we can alternatively keep the shape unchanged, \ie as in the original scene, via transforming the original shape with the predicted pose.

\subsubsection{3D scene generation and manipulation with the AtlasNet based model}

Figure \ref{fig:pcd_comparison} shows generation and manipulation of 3D scenes from scene graphs using the Graph-to-3D model together with AtlasNet~\cite{groueix2018atlas}. 
It can be noted that similarly to the DeepSDF~\cite{park2019deepsdf} based encodings (\cf~Figure~4 in the main paper), the model based on AtlasNet encodings is capable of generating correct point clouds under their diverse class categories, which are consistent with their semantic relationships. Further the manipulations are also appropriate with respect to changes in the graphs.

\begin{figure*}[t!]
    \centering
    \subfloat[DeepSDF encodings]{\includegraphics[width=\linewidth]{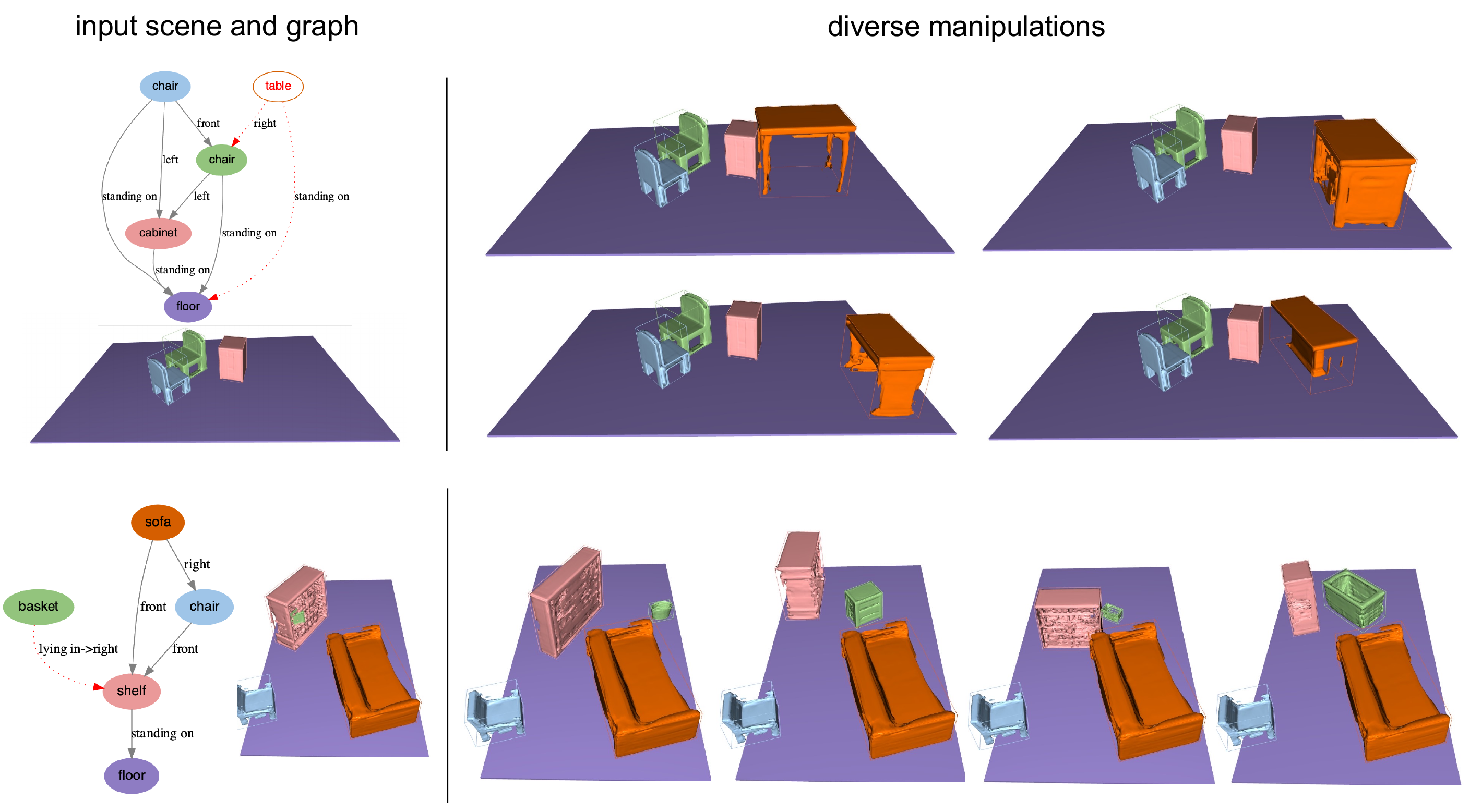}} \\
    \subfloat[AtlasNet encodings]{\includegraphics[width=0.9\linewidth]{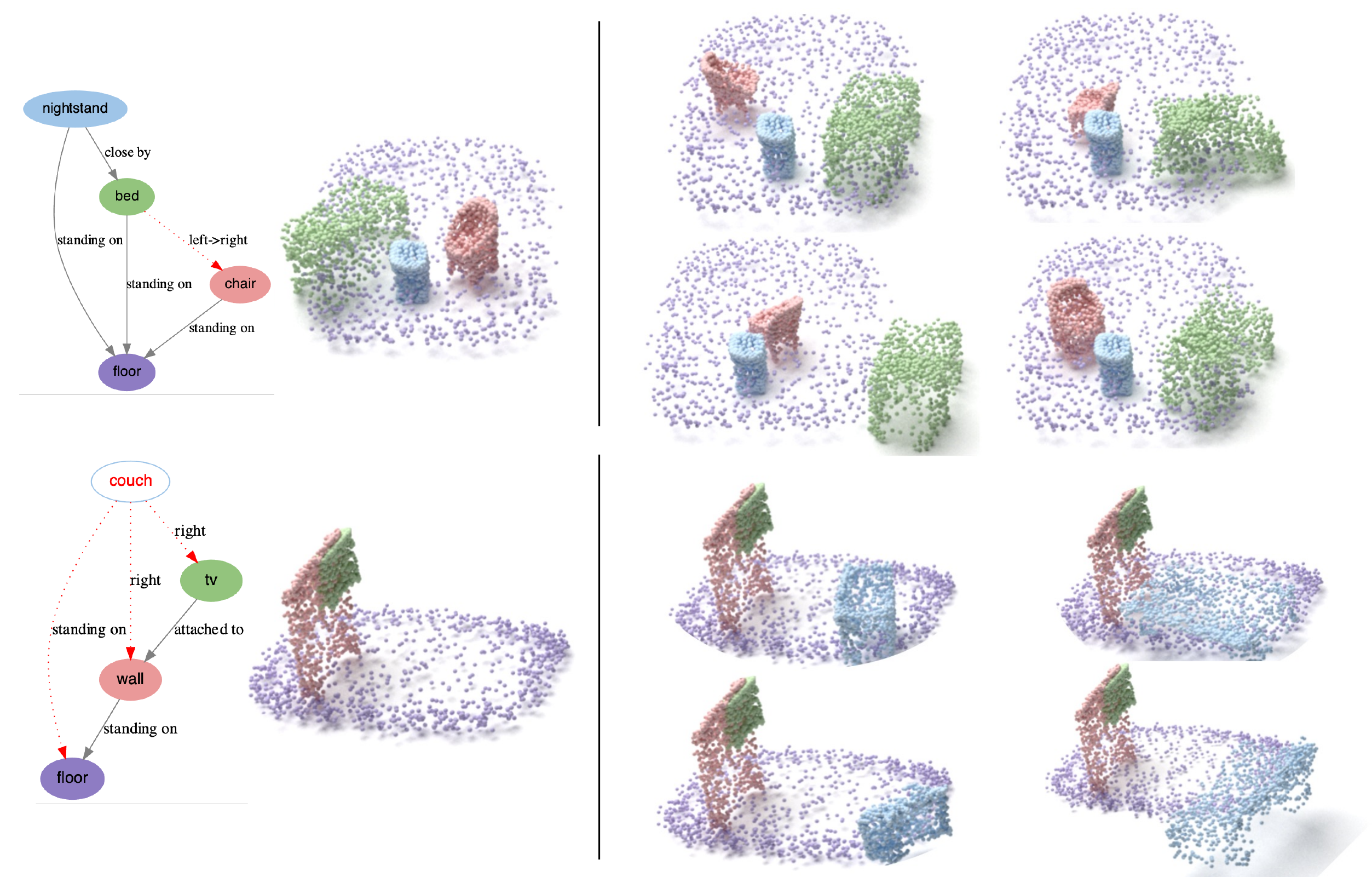}}
    \caption{Diverse generation of shapes and layout during manipulation. Given an input graph and correspondingly generated scene (left), we obtain diverse results (right) for the added or changed objects.}
    \label{fig:diversity}
\end{figure*}

\subsubsection{Diverse scene generation}
In Figure~\ref{fig:diversity} we want to demonstrate that Graph-to-3D is able to generate a diverse set of manipulations. To this end, we first generate a scene given only a semantic scene graph. Subsequently, we apply changes including additions and relationship changes to the graph and let the model repeatedly incorporate them. Notice that we run this experiment on top of both generative models, \ie AtlasNet and DeepSDF (\cf Figure~\ref{fig:diversity} a) and b)). Hence, for the same input, Graph-to-3D is capable of incorporating diverse manipulations in terms of both -- 3D shape as well as 3D location and orientation. 

\end{document}